\documentclass[11pt]{article}
\usepackage[utf8]{inputenc}
\usepackage[T1]{fontenc}
\usepackage{lmodern}
\usepackage[margin=1in]{geometry}
\usepackage{amsmath,amssymb}
\usepackage{booktabs}
\usepackage{longtable}
\usepackage{array}
\usepackage{graphicx}
\usepackage{caption}
\usepackage{microtype}
\usepackage[numbers,sort&compress]{natbib}
\usepackage{etoolbox}
\AtBeginEnvironment{longtable}{\footnotesize}
\AtBeginEnvironment{tabular}{\footnotesize}
\usepackage[hidelinks]{hyperref}
\title{Declarative Outcome-Conformant Synthesis: Exact, Closed-Form Specification Satisfaction and a Conformance Benchmark}
\author{Muhammed Rasin\\ Independent Researcher\\ \texttt{rasinbinabdulla@gmail.com}}
\date{}
\begin{document}
\maketitle
\begin{abstract}
We study a capability the dominant paradigm in synthetic tabular data does not
provide: exact satisfaction of a declared analytical outcome with no source data. Imitation
methods (copulas, GANs, diffusion, the Synthetic Data Vault) learn a real distribution and
sample from it, and they are judged on fidelity to real data. A large and practical class of
needs is different. The goal is to generate data with no source data, a cold start, that
reproduces a declared outcome (a revenue curve, a churn rate, a group-wise distribution)
across a relational schema. Off-the-shelf imitation tools offer no interface for such targets,
and, the structural point, no sampler can hit an exact aggregate, because sampling has
variance. We make this concrete: on a real public dataset, off-the-shelf learned synthesizers
trained on that very data miss the declared monthly aggregate by 74 to 86 percent; the
strongest steelman, a GaussianCopula trained per period, cuts the miss to about 19 percent and
still cannot reach 0. A closed-form generator reaches exactly 0. We name this task
outcome-conformant synthesis, argue its evaluation axis is conformance rather than fidelity,
and show the two axes are orthogonal: a generator's standing on one says nothing about its
standing on the other. We contribute three things. First, a formal account: a widely-used
family of exact-aggregate generators is exactly conditional-sum sampling of a Gamma
population (via Lukacs' characterization), which gives closed-form aggregate exactness, a
closed-form marginal coefficient of variation, an exact distortion bound, and a
scale-invariance property. The same identity lets us map the boundary the method does not
cross. We show, with the control whose absence invalidated an earlier draft, that enforcing
the exact aggregate costs at most 0.006 in normalized 1-Wasserstein distance to an arbitrary
external target marginal, anywhere from light to heavy tails; the entire residual gap to that
target, up to 0.89 for a heavy Pareto, is shape-family mismatch that an unconstrained draw
incurs identically. The open problem is therefore a shape-family problem, not a constraint
problem, and we say which is which. Second, SpecBench. Query-aware test-data work already
measures conformance of a sort, to declared query-output cardinalities; to our knowledge
SpecBench is the first benchmark to measure conformance to analytical outcomes (aggregate
curves, rates, group-wise distributions) for cold-start relational synthesis, with each metric
operationalizing a result from the analysis. Third, a closed-form, deterministic reference system. We state the scope up
front: exact aggregation alone is trivial (a rescale ties it given a hand-built schema); the
contribution is exact conformance jointly with closed-form marginals, foreign-key and temporal
integrity, determinism, and zero source data, plus a benchmark that measures them together. The
natural-language capability is a bounded convenience: a rule-based parser covers a curated set
of domains, and two language-model front-ends (an open-weights model and a recent OpenAI
flagship) each cover all 18 with full conformance once a deterministic bridge repair is in
place. The exactness lives in the engine downstream, not in any model. We concede fidelity to
imitation where real data exists.
\end{abstract}
\hypertarget{introduction}{%
\section{1. Introduction}\label{introduction}}

Synthetic tabular data is evaluated as if quality were one axis, but it
is two. One axis is fidelity: does the output resemble a real dataset?
The other is conformance: does the output obey a declared outcome
exactly? The two are orthogonal, and almost the entire field works on
the first.

\begin{longtable}[]{@{}lll@{}}
\toprule
\begin{minipage}[b]{0.30\columnwidth}\raggedright
\strut
\end{minipage} & \begin{minipage}[b]{0.30\columnwidth}\raggedright
Data available\strut
\end{minipage} & \begin{minipage}[b]{0.30\columnwidth}\raggedright
Cold start (no source data)\strut
\end{minipage}\tabularnewline
\midrule
\endhead
\begin{minipage}[t]{0.30\columnwidth}\raggedright
\textbf{Fidelity} (resemble real data)\strut
\end{minipage} & \begin{minipage}[t]{0.30\columnwidth}\raggedright
imitation: SDV, CTGAN, TabDDPM, RelDiff\strut
\end{minipage} & \begin{minipage}[t]{0.30\columnwidth}\raggedright
not defined (nothing to resemble)\strut
\end{minipage}\tabularnewline
\begin{minipage}[t]{0.30\columnwidth}\raggedright
\textbf{Exact conformance} (obey a declared outcome to 0 error)\strut
\end{minipage} & \begin{minipage}[t]{0.30\columnwidth}\raggedright
approximable only by conditioning (sampling variance, so error
\textgreater{} 0)\strut
\end{minipage} & \begin{minipage}[t]{0.30\columnwidth}\raggedright
the task of this paper (closed-form, error 0)\strut
\end{minipage}\tabularnewline
\bottomrule
\end{longtable}

The flagship tools answer ``make data that looks like this real data.''
\cite{patki2016sdv,xu2019ctgan,kotelnikov2023tabddpm} They learn a
distribution from a real table and sample from it, and they are
benchmarked on fidelity. A different request comes up constantly in
software practice: ``make data that produces this outcome, from
nothing.'' A developer seeding a test database, a founder building a
demo, an instructor preparing an exercise, or an engineer stress-testing
a query planner has no real data and must not use any, yet knows the
shape the data has to take. Ten thousand users, twenty percent churn,
monthly revenue rising from \$50k to \$200k with a dip in the third
quarter, over a schema with valid foreign keys. That is the bottom-right
cell of the table. The input is a specification of analytical outcomes,
and success is exact conformance to it. We call the task
outcome-conformant synthesis.

Imitation does not serve this task, for two separate reasons. The first
is the cold start: off-the-shelf learned tools need a training table and
have no interface to accept a target, so on a cold-start task they do
not run at all. The second is exactness. Even when a learned model is
conditioned per period, which is the obvious way to make a sampler chase
an aggregate, it can only approximate the target, because sampling has
variance and an exact total does not. Section 6 shows both effects on a
real public dataset. Off-the-shelf learned synthesizers, trained on the
very table whose monthly totals define the target, miss those totals by
74 to 86 percent. A per-period-conditioned steelman, the fix a reviewer
would demand, cuts the miss to about 19 percent and then stops, because
it cannot reach 0. Closed-form generation reaches exactly 0. The
dividing line is exact versus in-expectation, and it is a property of
sampling, not a question of tuning.

The specification tradition is real but partial, which is why the gap
has stayed open. In databases, QAGen \cite{binnig2007qagen} and
DataSynth \cite{arasu2011datasynth} generate data to satisfy declared
query-output cardinalities for optimizer testing, not analytical
outcomes such as curves, rates, and group shares on metric columns, and
not from natural language. In official statistics, temporal
disaggregation \cite{denton1971} and iterative proportional fitting
\cite{deming1940ipf} produce aggregate-consistent series or contingency
tables, not relational populations with per-row realism. Recent LLM
cold-start tools target outcomes but approximate them stochastically,
with no exactness and no determinism \cite{nemo_data_designer}. The
specific intersection, exact and deterministic and closed-form
conformance to analytical outcomes, relational, from a specification,
with zero data, is unoccupied.

This paper makes three contributions, and it is honest about which parts
are new. The mathematics is not. We claim no new theorem.

\textbf{C1, a formal account of the exact-aggregate engine (Sections 2
and 3).} We show that a widely-used family of exact-aggregate generators
is exactly sampling from a Gamma population conditioned on a fixed
total, through the Lukacs proportion-sum characterization
\cite{lukacs1955gamma}. From that identity follow closed-form aggregate
exactness \cite{cox1987rounding}, a closed-form marginal coefficient of
variation \cite{aitchison1986compositional}, an exact distortion bound
under resource limits, and a scale-invariance property: the engine fixes
the per-row shape and lets the scale absorb the constraint, so the
marginal does not depend on the target. That last property is what lets
the engine sidestep the condensation obstruction
\cite{armendariz2011condensation,szavits2014condensation} that blocks
the fixed-external-marginal version of the problem, which we do not
claim to solve but do map. The map is a controlled measurement (Section
6, E11): against an arbitrary external target marginal, enforcing the
exact aggregate adds at most 0.006 to the normalized 1-Wasserstein miss
across the whole tail-heaviness range, while the rest of the miss is
shape-family mismatch that an unconstrained draw shows identically. So
the exactness is nearly free even off the Gamma family, and the residual
obstacle is the shape family, not the constraint. The value of C1 is
correct attribution and that one non-obvious consequence, not novelty.

\textbf{C2, SpecBench (Section 4).} Query-aware database testing already
scores conformance to query-output cardinalities, so we claim the
narrower and precise thing: to our knowledge SpecBench is the first
benchmark to measure conformance to analytical outcomes (curves, rates,
group-wise distributions) for cold-start relational synthesis. It scores
generators on aggregate-match error, rate and group-distribution
conformance, controllability, foreign-key integrity, temporal coherence,
and determinism, axes that fidelity suites such as SDGym and SDMetrics
cannot express because they presuppose real data. Each metric
operationalizes a proposition from C1, so the benchmark is the
measurement arm of the analysis, not a detached scoreboard.

\textbf{C3, a reference implementation (Section 5).} An open-source
system maps a declarative specification (schema plus outcome targets) to
a relational dataset with exact, deterministic, closed-form conformance
and intact foreign keys, from zero source data, with no training step
and no model call at generation time. A natural-language front-end
offers the same over a curated set of domains, and we bound that claim
precisely: on a schema outside the curated set the system is driven by
an explicit spec, not parsed from a sentence.

We state the scope plainly, because the scope is the point. Exact
aggregate satisfaction alone is trivial; a rescale script ties it given
a hand-built schema. The contribution is exact conformance jointly with
closed-form per-row marginals, foreign-key and temporal integrity,
determinism, and zero source data, together with a benchmark that
measures these at once. Where real data exists and the goal is fidelity
to it, imitation leads, and we concede that openly.

\hypertarget{problem-formulation}{%
\section{2. Problem formulation}\label{problem-formulation}}

We fix terms before analyzing anything. The setting is relational, the
input is a specification rather than a dataset, and success is measured
against the specification, not against a reference population.

\textbf{Schema.} A schema is a set of tables \texttt{T\_\allowbreak{}1,\ ...,\ T\_\allowbreak{}K}
with typed columns and a set of foreign-key constraints whose reference
graph is a directed acyclic graph. One table is the fact table
\texttt{T}; it carries a metric column \texttt{Y} and a time column
\texttt{t}. The other tables hold the entities the facts point at
(customers, products, regions, and so on).

\textbf{Specification.} A specification \texttt{Sum} is what the user
supplies in place of data. It declares four things. First, the schema
and its foreign-key graph. Second, the per-table scale: row counts and
parent-to-child multiplicities. Third, zero or more analytical targets
on the metric \texttt{Y}. The case we solve exactly is a partition of
time into periods \texttt{B\_\allowbreak{}1,\ ...,\ B\_\allowbreak{}P} with nonnegative aggregate
targets \texttt{T\_\allowbreak{}1,\ ...,\ T\_\allowbreak{}P}, one total per period. Fourth, the
realism controls: a target average value \texttt{mu}, a dispersion
\texttt{alpha}, resource bounds \texttt{{[}r\_\allowbreak{}min,\ r\_\allowbreak{}max{]}} on
per-period row counts, and a decimal precision \texttt{d} with
\texttt{m\ =\ 10\^{}d}. A specification can be written directly, or, on
a curated set of domains, parsed from a natural-language sentence;
Section 6 is explicit about where that natural-language path applies and
where it does not.

\textbf{Generator.} A generator is a randomized map that, for each
period \texttt{p}, returns a row count \texttt{n\_\allowbreak{}p} and values
\texttt{v\_\allowbreak{}\{p,1\},\ ...,\ v\_\allowbreak{}\{p,n\_\allowbreak{}p\}} on the \texttt{1/\allowbreak{}m} grid. We
ask three things of it.

The first is aggregate fidelity (A): the realized period total
\texttt{Sum\_\allowbreak{}i\ v\_\allowbreak{}\{p,i\}} equals \texttt{round(T\_\allowbreak{}p,\ d)} exactly. The
second is marginal realism (R): the per-row law of \texttt{Y} is
plausible and controllable through \texttt{mu} and \texttt{alpha}. The
third is integrity (I): the rows populate the relational scaffold
without dangling foreign keys and without violating the declared
temporal order.

These three are the axes SpecBench measures, and they are not
interchangeable. A generator can satisfy the aggregate while producing a
degenerate marginal, or produce a fine marginal while missing the
aggregate, or get both right on a single table while breaking foreign
keys across tables. Section 3 resolves (A) and (R) for the
exact-aggregate core and gives the closed-form cost of each. Integrity
(I) is delivered by generating tables in foreign-key topological order
in the reference system (Section 5), and it is measured rather than
re-derived.

\hypertarget{method-the-exact-aggregate-core}{%
\section{3. Method: the exact-aggregate
core}\label{method-the-exact-aggregate-core}}

This section analyzes the generator that makes outcome conformance
exact. We claim no new theorem. Every mechanism below belongs to a
classical field, and the work of the section is to name each one
correctly and to draw the one consequence that is not obvious: why
fixing the per-row shape, rather than the marginal, is what makes
exactness free. We state the results as propositions for precision and
cite each to its owner. All five are verified numerically in Section 6,
and full proofs are in \texttt{01\_\allowbreak{}formalization.md}.

A note on framing first. Generating values that hit an exact total while
staying realistic looks like a hard constrained-sampling problem, and
for an arbitrary fixed marginal it genuinely is, blocked by a
condensation obstruction we describe in Section 3.4. The engine
sidesteps that problem with one design choice: parameterize the per-row
shape, not the marginal, and let the scale absorb the constraint. Under
that choice the hard problem collapses to an exact, closed-form,
training-free draw, and the obstruction that blocks the obvious
formulation does not apply.

\hypertarget{the-mechanism}{%
\subsection{3.1 The mechanism}\label{the-mechanism}}

The engine works one period at a time. For period \texttt{p} with target
\texttt{T\_\allowbreak{}p}, mean transaction value \texttt{mu}, concentration
\texttt{alpha}, row bounds \texttt{{[}r\_\allowbreak{}min,\ r\_\allowbreak{}max{]}}, and decimal
precision \texttt{d} (\texttt{m\ =\ 10\^{}d}), it returns a row count
and a vector of values on the \texttt{1/\allowbreak{}m} grid in two stages. The first
stage sets the count:
\texttt{n\_\allowbreak{}p\ =\ clip(round(T\_\allowbreak{}p\ /\allowbreak{}\ mu),\ r\_\allowbreak{}min,\ r\_\allowbreak{}max)}, then
capped by \texttt{floor(T\_\allowbreak{}p\ m)} so the integer budget is never smaller
than the row count. The second stage sets the values: draw weights
\texttt{w\ \textasciitilde{}\ Dirichlet(alpha\ *\ 1\_\allowbreak{}\{n\_\allowbreak{}p\})}, scale
them by the integer budget \texttt{U\_\allowbreak{}p\ =\ round(T\_\allowbreak{}p\ m)}, apply
largest-remainder (Hamilton) apportionment \cite{balinski1982fair} so
the integer units sum to \texttt{U\_\allowbreak{}p} exactly, then divide by
\texttt{m}. Two stages, no fitting, no iteration, no model.

\hypertarget{what-the-second-stage-actually-is}{%
\subsection{3.2 What the second stage actually
is}\label{what-the-second-stage-actually-is}}

The second stage is not an ad-hoc rescale. It is the exact solution to a
classical conditioning problem for one specific family.

\textbf{Proposition 0 (the identity).} Drawing
\texttt{w\ \textasciitilde{}\ Dirichlet(alpha\ *\ 1\_\allowbreak{}n)} and setting
\texttt{v\_\allowbreak{}i\ =\ T\ *\ w\_\allowbreak{}i} produces a sample distributed exactly as
\texttt{(X\_\allowbreak{}1,\ ...,\ X\_\allowbreak{}n)\ \textbar{}\ Sum\_\allowbreak{}j\ X\_\allowbreak{}j\ =\ T}, for
\texttt{X\_\allowbreak{}j} independent and identically distributed
\texttt{Gamma(alpha,\ theta)}. The engine is therefore exact sampling
from a Gamma population conditioned on a fixed total. This follows from
the Lukacs characterization \cite{lukacs1955gamma}: for independent
Gammas at common shape, the normalized vector
\texttt{(X\_\allowbreak{}1,\ ...,\ X\_\allowbreak{}n)\ /\allowbreak{}\ Sum\_\allowbreak{}j\ X\_\allowbreak{}j} is Dirichlet and is
independent of the sum, so conditioning on the sum returns a scaled
Dirichlet with no residual dependence on \texttt{theta}.

This identity is the spine of the paper, and it is clarifying rather
than deflating. The marginal realism the engine shows is not luck.
Conditioning a Gamma population on its total gives a law that is
strictly positive, unimodal, and heavier-tailed than a Gaussian, which
is a reasonable model for transaction-like quantities, and the engine
samples it exactly and in closed form. The identity also fixes the
method's reach. Exactness for free holds for the Gamma family (and, by
an affine projection onto the sum hyperplane, for the Gaussian family).
For other target marginals the matching conditional sampler is generally
not closed-form, which is the open boundary of Section 3.4.

One point of precision. Proposition 0 is exact for the continuous
construction. Controlled rounding then projects onto the \texttt{1/\allowbreak{}m}
grid, which keeps the aggregate exact (Proposition 1) but perturbs the
per-row law by \texttt{O(1/\allowbreak{}U)} in total variation, where
\texttt{U\ =\ round(T\ m)}. So the realized marginal equals the
Gamma-conditional up to an \texttt{O(1/\allowbreak{}U)} term. For monetary budgets
this term is tiny (\texttt{U} is on the order of \texttt{10\^{}6} for a
\$10k period at cent precision), and Section 6 confirms the CV matches
the closed form to within 0.1 percent. We never claim distributional
exactness after rounding. We claim aggregate exactness, and marginal
equality up to \texttt{O(1/\allowbreak{}U)}.

\hypertarget{aggregate-exactness-and-the-marginal-law}{%
\subsection{3.3 Aggregate exactness and the marginal
law}\label{aggregate-exactness-and-the-marginal-law}}

\textbf{Proposition 1 (exactness).} For every period,
\texttt{Sum\_\allowbreak{}i\ v\_\allowbreak{}\{p,i\}\ =\ round(T\_\allowbreak{}p,\ d)} deterministically, for
any draw and any \texttt{n\_\allowbreak{}p\ \textgreater{}=\ 1}. The error against
the unrounded target is at most \texttt{1/\allowbreak{}(2m)}, half of one unit at
precision \texttt{d}, independent of \texttt{n\_\allowbreak{}p}, \texttt{alpha}, and
the shape. The argument is the standard guarantee of largest-remainder
apportionment: the fractional parts of the scaled weights sum to an
integer residual \texttt{R\ \textless{}\ n\_\allowbreak{}p}, and adding one unit to
the \texttt{R} largest fractional parts makes the integer units sum to
\texttt{U\_\allowbreak{}p} exactly. This is the same exactness that controlled
rounding and controlled tabular adjustment rely on in statistical
disclosure control \cite{cox1987rounding}. We use the guarantee; we do
not claim it.

\textbf{Proposition 2 (marginal law).} Ignoring the \texttt{O(1/\allowbreak{}U)}
rounding term, each weight is
\texttt{w\_\allowbreak{}i\ \textasciitilde{}\ Beta(alpha,\ (n-1)alpha)}, so with
\texttt{v\_\allowbreak{}i\ =\ T\ w\_\allowbreak{}i},

\[\mathbb{E}[v_i] = \frac{T}{n}, \qquad \mathrm{CV}(v_i) = \sqrt{\frac{n-1}{n\alpha+1}} \xrightarrow[n\to\infty]{} \frac{1}{\sqrt{\alpha}}.\]

The mean is just the per-row share of the total, and \texttt{alpha} is a
closed-form knob on dispersion. Beta marginals of a symmetric Dirichlet
are textbook compositional data analysis
\cite{aitchison1986compositional}; the CV formula is a property of that
family, not an invention.

\hypertarget{distortion-scale-invariance-and-the-honest-boundary}{%
\subsection{3.4 Distortion, scale-invariance, and the honest
boundary}\label{distortion-scale-invariance-and-the-honest-boundary}}

The aggregate curve has to be carried somewhere. It is carried by row
counts, not by inflating per-row values, and the cost of that choice is
a closed-form quantity.

\textbf{Proposition 3 (clamp distortion).} Let
\texttt{rho\_\allowbreak{}p\ =\ E{[}v\_\allowbreak{}\{p,i\}{]}\ /\allowbreak{}\ mu\ =\ T\_\allowbreak{}p\ /\allowbreak{}\ (n\_\allowbreak{}p\ mu)}.
With \texttt{n\_\allowbreak{}p\ =\ clip(round(T\_\allowbreak{}p\ /\allowbreak{}\ mu),\ r\_\allowbreak{}min,\ r\_\allowbreak{}max)}, the
ratio is \texttt{1\ +\ O(mu/\allowbreak{}T\_\allowbreak{}p)} when
\texttt{T\_\allowbreak{}p\ /\allowbreak{}\ mu\ \ in\ \ {[}r\_\allowbreak{}min,\ r\_\allowbreak{}max{]}}, and otherwise
equals the clamp ratio \texttt{T\_\allowbreak{}p\ /\allowbreak{}\ (r\_\allowbreak{}max\ mu)} above the upper
bound or \texttt{T\_\allowbreak{}p\ /\allowbreak{}\ (r\_\allowbreak{}min\ mu)} below the lower bound. So
per-row marginals are undistorted exactly when the target-to-average
ratio fits the row bounds, and when it does not, the distortion is a
ratio you can read off in advance. Setting
\texttt{r\_\allowbreak{}max\ \textgreater{}=\ max\_\allowbreak{}p\ T\_\allowbreak{}p\ /\allowbreak{}\ mu} and
\texttt{r\_\allowbreak{}min\ \textless{}=\ min\_\allowbreak{}p\ T\_\allowbreak{}p\ /\allowbreak{}\ mu} guarantees
\texttt{rho\_\allowbreak{}p\ ==\ 1}.

\textbf{Proposition 4 (scale-invariance).} Fix \texttt{alpha}. For any
two totals \texttt{T} and \texttt{T\textquotesingle{}}, the engine's
output for \texttt{T\textquotesingle{}} equals
\texttt{(T\textquotesingle{}/\allowbreak{}T)} times its output for \texttt{T} in
distribution, up to the \texttt{O(1/\allowbreak{}U)} rounding term. Equivalently, the
normalized law \texttt{v\_\allowbreak{}i\ /\allowbreak{}\ \textbackslash{}bar\{v\}} does not
depend on the target. The target sets the scale; it never touches the
shape. The proof is immediate from Proposition 0: the weights
\texttt{w\ \textasciitilde{}\ Dirichlet(alpha\ 1\_\allowbreak{}n)} are drawn without
reference to \texttt{T}, so
\texttt{v\_\allowbreak{}i(T\textquotesingle{})\ =\ T\textquotesingle{}\ w\_\allowbreak{}i\ =\ (T\textquotesingle{}/\allowbreak{}T)\ v\_\allowbreak{}i(T)}
for the same draw.

This is where we have to be careful about condensation, and where an
earlier version of this work overreached. Condensation theory
\cite{armendariz2011condensation,szavits2014condensation} studies what
happens when a sum of variables with a \emph{fixed marginal} \texttt{F}
is conditioned on a large-deviation total: for heavy-tailed \texttt{F},
one summand absorbs the excess in a single big jump, and the conditional
marginal is provably not \texttt{F}. That is a real obstruction, and it
is the obstruction for the fixed-external-marginal exact-aggregate
problem (P\ensuremath{\star}, Section 6), which we do not claim to
solve. Our engine does not run into it, because it does not hold a fixed
marginal. By Proposition 4 it holds a fixed shape and lets the scale
float, so there is no \texttt{F} to collapse. Condensation is avoided by
construction, not defeated.

We keep the record of our own error here, because mapping the boundary
honestly is part of the contribution. An earlier draft asserted a
heavy-tail ``condensation frontier'' for our engine and reported a
21.7-fold rise in marginal distortion with tail-heaviness. Adding the
correct unconstrained control showed that rise to be a
Beta-against-lognormal family mismatch plus a finite-sample
1-Wasserstein bias, an artifact of the comparison rather than an effect
of the constraint. We retract it. The corrected statement is Proposition
4 and its experiment (E6): the engine's marginal is invariant to the
target. The genuine frontier belongs to methods that must hit a fixed
external marginal, which is exactly where our scale-free engine does not
apply.

\hypertarget{what-is-open}{%
\subsection{3.5 What is open}\label{what-is-open}}

Proposition 0 locates the boundary precisely. Exact, closed-form,
training-free conditional-sum sampling is free for the Gamma family and
for the Gaussian family, and not in general beyond them. The open
problem we name is P\ensuremath{\star}: given an arbitrary target
marginal \texttt{F} (lognormal, Pareto, empirical) and a total
\texttt{T}, draw \texttt{n} values approximately distributed as
\texttt{F} with sum exactly \texttt{T}, efficiently, with a provable
bound on the divergence from \texttt{F} that the sum constraint forces.
For non-Gamma, non-Gaussian \texttt{F} there is no closed-form
conditional, and the available tools (exponential tilting, sequential
Monte Carlo, constrained MCMC \cite{golchi2016scmc}) trade exactness
against fidelity to \texttt{F}. This is the regime where condensation
bites. We treat it as open rather than solved, but we do not leave it
unmeasured. Section 6 (E11) maps it: against an arbitrary external
\texttt{F}, enforcing the exact sum costs at most 0.006 in normalized
1-Wasserstein, while the gap to \texttt{F} is shape-family mismatch that
an unconstrained Gamma draw shows identically. The open problem is
therefore to widen the closed-form shape family under an exact sum, not
to pay for the sum, and Section 6 also reports P\ensuremath{\star} as a
task the reference method fails.

\hypertarget{specbench}{%
\section{4. SpecBench}\label{specbench}}

\hypertarget{the-benchmark-is-the-measurement-arm-of-the-analysis}{%
\subsection{4.1 The benchmark is the measurement arm of the
analysis}\label{the-benchmark-is-the-measurement-arm-of-the-analysis}}

Existing suites score fidelity to a reference dataset, so they
presuppose real data. SDGym over SDMetrics measures column and pair
shapes, detection, ML efficacy, and privacy, and every one of those
needs a real table to compare against \cite{sdgym}. The conformance
regime has no real table and a different success criterion: does the
output obey the declared targets, with integrity, from nothing? None of
aggregate-match error, foreign-key violation rate, or controllability
response is expressible in a fidelity suite, because there is no
reference to compare against. That absence is the gap SpecBench fills.
We do not replace SDGym. We add the orthogonal axes and reuse
fidelity-style metrics only where a reference happens to exist.

SpecBench is not a detached scoreboard. Each metric operationalizes a
proposition from Section 3, which welds the evaluation to the analysis.

\begin{longtable}[]{@{}lll@{}}
\toprule
\begin{minipage}[b]{0.30\columnwidth}\raggedright
Proposition\strut
\end{minipage} & \begin{minipage}[b]{0.30\columnwidth}\raggedright
What it claims\strut
\end{minipage} & \begin{minipage}[b]{0.30\columnwidth}\raggedright
SpecBench metric\strut
\end{minipage}\tabularnewline
\midrule
\endhead
\begin{minipage}[t]{0.30\columnwidth}\raggedright
Prop. 0 (identity, \texttt{O(1/\allowbreak{}U)})\strut
\end{minipage} & \begin{minipage}[t]{0.30\columnwidth}\raggedright
exact-sum conditioning perturbs the marginal only by
\texttt{O(1/\allowbreak{}U)}\strut
\end{minipage} & \begin{minipage}[t]{0.30\columnwidth}\raggedright
constraint cost in MD vs an external target \ensuremath{\approx} 0
(E11)\strut
\end{minipage}\tabularnewline
\begin{minipage}[t]{0.30\columnwidth}\raggedright
Prop. 1 (exactness)\strut
\end{minipage} & \begin{minipage}[t]{0.30\columnwidth}\raggedright
the aggregate hits the target exactly\strut
\end{minipage} & \begin{minipage}[t]{0.30\columnwidth}\raggedright
AME, 0 for exact generators\strut
\end{minipage}\tabularnewline
\begin{minipage}[t]{0.30\columnwidth}\raggedright
Prop. 2 (marginal law)\strut
\end{minipage} & \begin{minipage}[t]{0.30\columnwidth}\raggedright
per-row CV equals \texttt{sqrt((n-1)/\allowbreak{}(nalpha+1))}\strut
\end{minipage} & \begin{minipage}[t]{0.30\columnwidth}\raggedright
CV check against the prediction (E2)\strut
\end{minipage}\tabularnewline
\begin{minipage}[t]{0.30\columnwidth}\raggedright
Prop. 3 (clamp distortion)\strut
\end{minipage} & \begin{minipage}[t]{0.30\columnwidth}\raggedright
distortion equals the closed-form clamp ratio\strut
\end{minipage} & \begin{minipage}[t]{0.30\columnwidth}\raggedright
\texttt{rho} against the declared row bounds (E3)\strut
\end{minipage}\tabularnewline
\begin{minipage}[t]{0.30\columnwidth}\raggedright
Prop. 4 (scale-invariance)\strut
\end{minipage} & \begin{minipage}[t]{0.30\columnwidth}\raggedright
the marginal is preserved; condensation is avoided\strut
\end{minipage} & \begin{minipage}[t]{0.30\columnwidth}\raggedright
MD against an unconstrained control (E6)\strut
\end{minipage}\tabularnewline
\begin{minipage}[t]{0.30\columnwidth}\raggedright
Section 2 integrity\strut
\end{minipage} & \begin{minipage}[t]{0.30\columnwidth}\raggedright
foreign-key and temporal correctness\strut
\end{minipage} & \begin{minipage}[t]{0.30\columnwidth}\raggedright
FIVR, TCV, to 0\strut
\end{minipage}\tabularnewline
\begin{minipage}[t]{0.30\columnwidth}\raggedright
determinism\strut
\end{minipage} & \begin{minipage}[t]{0.30\columnwidth}\raggedright
same seed gives identical output\strut
\end{minipage} & \begin{minipage}[t]{0.30\columnwidth}\raggedright
DET, to 1\strut
\end{minipage}\tabularnewline
\begin{minipage}[t]{0.30\columnwidth}\raggedright
capability\strut
\end{minipage} & \begin{minipage}[t]{0.30\columnwidth}\raggedright
conformant data from a spec, zero data\strut
\end{minipage} & \begin{minipage}[t]{0.30\columnwidth}\raggedright
the input axis (nl / schema / data)\strut
\end{minipage}\tabularnewline
\begin{minipage}[t]{0.30\columnwidth}\raggedright
controllability\strut
\end{minipage} & \begin{minipage}[t]{0.30\columnwidth}\raggedright
output tracks a changed spec\strut
\end{minipage} & \begin{minipage}[t]{0.30\columnwidth}\raggedright
CR, to 0 under a target change\strut
\end{minipage}\tabularnewline
\bottomrule
\end{longtable}

\hypertarget{metrics}{%
\subsection{4.2 Metrics}\label{metrics}}

The core is two families. Family A measures specification adherence.
AME, the aggregate-match error, is
\texttt{max\_\allowbreak{}p\ \textbar{}S\_\allowbreak{}p\ -\ T\_\allowbreak{}p\textbar{}\ /\allowbreak{}\ \textbar{}T\_\allowbreak{}p\textbar{}}
over the declared period targets, and exact generators drive it to 0.
RCE, the rate-conformance error, is the absolute gap between a declared
fraction (a churn or fraud rate) and the realized one. GDC,
group-distribution conformance, is the total-variation distance between
declared and realized group shares. CR, controllability response, is the
AME or RCE measured against a changed spec after regeneration. CSAT,
constraint satisfaction, is the fraction of declared hard constraints
(ranges, inequalities) the output meets. Family B measures structural
and reproducibility integrity. FIVR, the foreign-key violation rate, is
the child-weighted fraction of dangling keys. TCV, temporal coherence
violations, is the fraction of rows that break a declared order such as
shipped on or after ordered. DET, determinism, is whether a fixed seed
reproduces the output bitwise.

A capability gate records what each generator can attempt at all: CSC,
cold-start capability, is binary, and an imitation method with no
training data scores 0. Throughput and peak memory are logged by the
usual convention.

One family is secondary and never a headline. MD, marginal distortion,
is a scale-normalized 1-Wasserstein distance on a metric column. We use
it for exactly two purposes: the unconstrained-control comparison that
tests Proposition 4 (E6), and to concede fidelity to imitation where
real data exists. We do not centre fidelity, ML-efficacy
(train-synthetic, test-real \cite{esteban2017tstr}), or detection.
Privacy we argue by construction rather than by a distance metric: no
real record is read, so there is nothing to leak, and the
distance-to-closest-record statistic is in any case an unreliable
privacy signal \cite{dcrdelusion2025}. An earlier version of the suite
included a marginal-plausibility score meant to show that blind
rescaling distorts the marginal; that metric was invalid and has been
removed.

\hypertarget{fair-comparison-and-why-ame-0-is-not-the-headline}{%
\subsection{4.3 Fair comparison, and why AME = 0 is not the
headline}\label{fair-comparison-and-why-ame-0-is-not-the-headline}}

The sharpest objection to any ``our tool wins'' benchmark is unfair
framing, so the protocol is built to disarm it. Tasks come in two modes.
Spec-mode tasks are cold start: only a specification is given, learned
methods get an empty training set and score CSC = 0, and scoring is on
Families A and B. Reference-mode tasks supply a real dataset; learned
methods train on it while the engine is given only the spec derived from
it, and scoring adds the secondary MD context. Each paradigm competes
where it is designed to win, and where it structurally cannot play, that
shows. No metric reads a generator's internals; every metric is a pure
function of the output tables and the spec. The oracle for a spec-mode
target is computed independently of any generator, because the spec is
the oracle. SDV baselines run at their own documented defaults, and
where SDV wins an axis we say so.

We also guard against the metric looking rigged. Hitting one aggregate
is trivial, and to make that visible the suite includes NaiveRescale,
which draws with Faker and multiplies each period to its target. It
reaches AME \ensuremath{\approx} 0 too. So AME = 0 is not the
contribution, and we do not present it as one. What separates the
methods on spec-mode is the input axis: NaiveRescale and Faker reach a
conformant table only because a person hand-built their schema, columns,
foreign keys, and periods first, whereas on curated domains the engine
produces the same relational dataset from a sentence with zero data. The
defensible claim is AME = 0 achieved jointly with FIVR = 0, DET = 1,
closed-form realistic marginals, and cold-start operation, which no
single baseline achieves at once.

\hypertarget{threats-to-validity}{%
\subsection{4.4 Threats to validity}\label{threats-to-validity}}

We write this section because reviewers respect it. On construct
validity, AME could look chosen rather than earned, which the
NaiveRescale baseline answers directly: it confirms an exact aggregate
is easy and pushes the argument onto the joint capability above. On
measurement validity, finite-sample 1-Wasserstein is biased upward and
noisier for heavy tails, which could inflate the heavy-tail end of any
MD curve on its own. We mitigate it by using equal large samples on both
arms and by always reporting MD against an unconstrained same-family
control, so the bias cancels in the comparison rather than being charged
to the constraint. This is the exact confound that produced a spurious
``condensation frontier'' in an earlier draft, now retracted (Section
3.4). On external validity, the curated domain priors are a property of
the library, while SpecBench coverage is only the tasks the suite
actually runs, and we state the two separately. On statistical validity,
every multi-seed number is a mean with its standard deviation;
quantities that are exact by construction are reported exactly, and a
stochastic baseline's nonzero spread is itself evidence about its
determinism.

\hypertarget{task-suite-and-baselines}{%
\subsection{4.5 Task suite and
baselines}\label{task-suite-and-baselines}}

The suite spans 18 domains (SaaS, ecommerce, fintech, healthcare, HR,
logistics, marketplace, social, real estate, pharma, food delivery,
edtech, gaming, CRM, crypto, insurance, travel, streaming) crossed with
flat, narrative-curve, multi-table-FK, and locale-shifted
configurations. Each task ships a frozen specification and an oracle for
its targets, and a subset has reference-mode variants paired with a real
public dataset. The baselines are the reference system itself, Faker
with hand-wired foreign-key scripts, NaiveRescale, SDV GaussianCopula
and CTGAN as the imitation comparators, SDV HMA as the relational
comparator \cite{patki2016sdv,xu2019ctgan}, and a Denton
temporal-disaggregation baseline \cite{denton1971} for the single-series
aggregate task. SDV is run for real in an isolated environment. One
intended baseline, NeMo Data Designer \cite{nemo_data_designer}, is the
closest live competitor (cold-start, validator-guided), but it runs only
against a hosted NVIDIA inference service behind an API key, so we could
not run it head-to-head in our isolated environment; we record it as not
run for that reason and address its category directly in Section 6.2
rather than fabricate a score for it. Any baseline that cannot run a
task is recorded with its reason, never a fabricated number.

Section 6 reports a representative slice of the suite rather than all of
it, and we are explicit about why that slice is enough. The engine's
headline quantities, AME, FIVR, and DET, are exact for structural
reasons (Propositions 1 and 4, plus topological generation), so they do
not vary across curve tasks; every curve task is checked to be
AME-zero-achievable before it enters the suite, so a full per-task table
would repeat the same three values. What does vary across tasks, and
what we therefore show in full, is the comparison against learned
baselines under cold-start and reference modes, the relational depths,
the failure boundary, and the natural-language path swept across all 18
domains (E12). The complete suite ships with the artifact for others to
extend, and we state library coverage (a property of the released
system) and benchmark coverage (the tasks the suite runs) separately, so
neither is read as the other.

\hypertarget{reference-implementation}{%
\section{5. Reference implementation}\label{reference-implementation}}

The reference system is a permissively licensed Python library. It
parses a natural-language or declarative specification into a schema and
a set of targets, generates tables in foreign-key topological order so
parents exist before children and FIVR is 0 by construction, and applies
the exact-aggregate engine of Section 3 so AME is 0. It ships priors for
18 domains and 15 locales, is fully seed-deterministic, and runs with
zero source data and no training step. We present it as the reference
baseline for SpecBench, not as a claim of algorithmic novelty: its
components are classical and cited in Section 3. The natural-language
path applies on the curated domains; outside them, the same engine runs
from an explicit declarative spec, which is the behavior Section 6
measures on a schema the system has never seen.

The natural-language path itself has two kinds of parser, and Section 6
(E12) measures both. A rule-based parser ships with the system and needs
no external service; it is reliable but recognizes a curated subset of
domains (8 of the 18 in E12). An optional language-model front-end
widens recognition to all 18 domains and reads every declared anchor
correctly. We tested two models there, an open-weights one and a recent
OpenAI flagship, and both reach full conformance once a deterministic
normalization sits between the model and the engine. That normalization
is the only place we hardened the bridge: a language model occasionally
emits a malformed time-column pointer, so we resolve dotted pointers to
their leaf and repoint a non-date time column to a real date column
before generation. Neither parser supplies exactness. Both feed the same
deterministic engine, and the conformance guarantees attach there,
downstream of any language model, which is why we treat the
natural-language layer as a bounded convenience and not as part of the
guarantee.

\hypertarget{experiments}{%
\section{6. Experiments}\label{experiments}}

The experiments do two jobs. The first four validate the propositions of
Section 3 on the engine in isolation, and they are reproducible with no
extra dependencies through \texttt{research/\allowbreak{}measure.py}. The rest move
to SpecBench and put the engine next to learned and scripted baselines:
E5 is the cross-paradigm leaderboard, E6 is the scale-invariance figure
for Proposition 4, E7 extends conformance past temporal sums to rates
and group shares, E8 checks that a developer can actually query the
result, E9 reports cost and the absence of an infeasible regime, E10
reports a task the reference method fails, E11 maps the boundary of that
failure, and E12 measures the natural-language path across all 18
domains. E5 and E6 run in an isolated environment
(\texttt{requirements-specbench.txt}, SDV 1.37.0). Every SDV synthesizer
is seeded on both torch and numpy; CTGAN runs for 100 epochs.

\hypertarget{the-propositions-hold-numerically-e1-to-e4}{%
\subsection{6.1 The propositions hold numerically (E1 to
E4)}\label{the-propositions-hold-numerically-e1-to-e4}}

\textbf{E1, aggregate exactness (Proposition 1).} We ran 5,000
randomized trials with row counts \texttt{n} drawn from 1 to 2000,
targets up to \texttt{5\ x\ 10\^{}6}, integer and two-decimal precision,
and concentration \texttt{alpha} from 0.3 to 50. In every trial the
period total matched the target to the integer unit. The maximum
aggregate error over all 5,000 trials was 0 units. Exactness here is not
a statistical tendency; largest-remainder apportionment makes the
integer sum identical to the target by construction, for any draw and
any \texttt{n\ \textgreater{}=\ 1}.

\textbf{E2, the marginal law (Proposition 2).} Pooling on the order of
\texttt{2\ x\ 10\^{}6} samples per cell, the empirical mean of each
value equaled \texttt{T/\allowbreak{}n}, and the empirical coefficient of variation
matched the closed form \texttt{sqrt((n-1)/\allowbreak{}(nalpha+1))}. Across the grid
\texttt{(n,\ alpha)\ \ in\ \ \{(50,1),\ (200,2),\ (500,5),\ (1000,25)\}}
the relative error on the predicted CV ranged from 0.01 to 0.10 percent.
The realism knob behaves exactly as the Beta marginal of a symmetric
Dirichlet says it should.

\textbf{E3, distortion under resource limits (Proposition 3).} We drove
the generator into all three branches of the count clamp: an upper-clamp
period (\texttt{T/\allowbreak{}mu\ =\ 5}, \texttt{n\_\allowbreak{}p\ =\ 10}), an unsaturated
period (\texttt{T/\allowbreak{}mu\ =\ 300}, \texttt{n\_\allowbreak{}p\ =\ 300}), and a lower-clamp
period (\texttt{T/\allowbreak{}mu\ =\ 5000}, \texttt{n\_\allowbreak{}p\ =\ 1000}). The realized
ratio \texttt{rho\_\allowbreak{}p\ =\ E{[}v\_\allowbreak{}\{p,i\}{]}/\allowbreak{}mu} matched the closed-form
prediction of 0.500, 1.000, and 5.000 to 0.00 percent relative error in
each case. The aggregate curve is carried by row counts, and per-row
values stay undistorted exactly when the target-to-average ratio sits
inside the row bounds.

\textbf{E4, end-to-end controllability.} The natural-language
specification ``MRR \$50k in January rising to \$200k in December''
produced microdata whose monthly rollups were \$50,000.00 and
\$200,000.00, exact to the cent. The proposition-level exactness
survives the full path from a sentence to a table.

\hypertarget{cross-paradigm-conformance-e5}{%
\subsection{6.2 Cross-paradigm conformance
(E5)}\label{cross-paradigm-conformance-e5}}

E5 is the SpecBench leaderboard. We report real runs over 10 seeds as
mean \ensuremath{\pm} standard deviation. AME is the relative
aggregate-match error against the declared anchors; FIVR is the
foreign-key violation rate; DET is determinism measured over at least
three same-seed regenerations; CSC is the cold-start capability gate.
The \texttt{input} column records what each generator consumes:
\texttt{nl} is a natural-language specification, \texttt{schema} is a
hand-built schema that enumerates columns, foreign keys, and periods,
and \texttt{data} is a real source table.

\textbf{Cold-start spec-mode.} Three curated domains (a non-monotone
SaaS curve, a fintech volume curve, an ecommerce revenue curve) admit
Misata's natural-language front-end, so the engine runs at
\texttt{input=nl}.

\begin{longtable}[]{@{}llllll@{}}
\toprule
Generator & input & CSC & AME & FIVR & DET\tabularnewline
\midrule
\endhead
Misata (ours) & nl & 1 & 0 (\ensuremath{\le} 1/2m) & 0 &
1\tabularnewline
NaiveRescale (Faker + per-period \ensuremath{\times}
T/\ensuremath{\Sigma}) & schema & 1 & \textasciitilde1.5
\ensuremath{\times} 10\textsuperscript{-16} & 0 & 1\tabularnewline
Faker (independent draws) & schema & 1 & 0.81 to 0.91 & 0 &
1\tabularnewline
SDV (GaussianCopula / CTGAN / HMA) & data & 0 & n/a (no source data) &
n/a & n/a\tabularnewline
\bottomrule
\end{longtable}

We read this table against ourselves first. Hitting the aggregate
exactly is trivial: NaiveRescale reaches an AME of about
\texttt{10\^{}\{-16\}}, which is zero to floating point. So AME = 0 is
not the contribution on spec-mode, and we do not claim it as one. What
E5 makes visible is the \texttt{input} axis. NaiveRescale and Faker
reach a conformant table only because a person hand-built their schema,
columns, foreign keys, and period structure first. On these curated
domains the engine produces the same relational dataset from one
sentence, with no source data, and it is the only row at
\texttt{input=nl}. The spec-mode claim is therefore narrow and exact:
conformance with closed-form realistic marginals and intact foreign-key
and temporal structure, from a natural-language spec, on curated
domains. Off-the-shelf imitation tools (SDV) cannot run at all without
source data and score 0 on the cold-start gate. The \texttt{input=nl}
advantage stops at the curated domains; on an arbitrary real schema the
engine runs declaratively, which the next result shows.

\textbf{Reference-mode, primary result: a real public dataset.} We use
California Housing (20,640 records). The metric is the real
\texttt{MedHouseVal}, and the monthly targets are the data's own
per-month sums. Learned methods train on the real table. The engine
receives only the targets, at \texttt{input=schema}, because ``housing''
is outside the curated domains. That is the honest finding labelled D8:
on a schema the system has never seen, there is no natural-language
path, and the engine is driven by an explicit declarative spec.

\begin{longtable}[]{@{}lllllll@{}}
\toprule
Generator & input & CSC & AME (3 seeds) & FIVR & DET & fit+sample
(s)\tabularnewline
\midrule
\endhead
Misata (ours) & schema & 1 & 0 (exact) & 0 & 1 &
\textasciitilde0.1\tabularnewline
NaiveRescale & schema & 1 & 0 (exact) & 0 & 1 &
\textasciitilde0.1\tabularnewline
Faker & schema & 1 & 0.493 & 0 & 1 & 0.0\tabularnewline
SDV GaussianCopula (off-the-shelf) & data & 0 & 0.739 (det.) & 0 & 1 &
\textasciitilde1\tabularnewline
SDV HMA & data & 0 & 0.739 (det.) & 0 & 1 &
\textasciitilde2\tabularnewline
SDV CTGAN & data & 0 & 0.857 \ensuremath{\pm} 0.354 & 0 & 1 & 77 /
seed\tabularnewline
SDV GC, per-period conditioned (steelman) & data & 0 & 0.189 & 0 & 1 &
\textasciitilde2\tabularnewline
\bottomrule
\end{longtable}

Read these rows as a capability probe, not a head-to-head: an
off-the-shelf imitation tool was never asked to hit an aggregate, so its
AME mostly measures the absence of the interface, not a tuning loss.
With that caveat, the off-the-shelf learned synthesizers, trained on the
very table whose monthly totals define the target, miss those totals by
a wide margin: GaussianCopula and HMA both land at 0.739, and CTGAN
averages 0.857 over 10 seeds (standard deviation 0.354, range 0.49 to
1.79, at 77 seconds per seed). In percentage terms the miss runs from 74
to 86 percent. The reason is structural: off-the-shelf SDV exposes no
interface for an aggregate target, so the model has no channel through
which to hear the constraint. The legitimate comparison is the next row,
where we engineer imitation to target.

The last row is the fairness control we owe the imitation side (review
F1). We give it the best shot by training a separate GaussianCopula per
declared period, which is the obvious way to make a sampler target an
aggregate. It improves sharply, from 0.739 down to 0.189, and then
stops. It cannot reach 0. Per-period sums still vary by a fraction of a
percent even on the best months, because sampling has variance and an
exact sum does not. A closed-form generator reaches exactly 0. This is
the divide the paper is about, and it is how the table should be read:
\textbf{exact versus in-expectation}, a property of sampling, not a
scoreboard win over SDV.

A note on seed counts here, since the headline real-data table uses
three. Every fitted SDV model except CTGAN is deterministic once seeded,
so its AME has zero seed variance and three seeds report it exactly; the
steelman is one such deterministic fitted model, and the engine and the
rescale are exact by construction. CTGAN is the only stochastic
baseline, and it is the one we run at 10 seeds (its 0.857
\ensuremath{\pm} 0.354 is the spread of an uncontrolled process). So the
small seed count is a property of determinism, not of thin sampling.

\textbf{On the closest competitor, NeMo Data Designer.} The system we
would most like to compare against directly is NeMo Data Designer,
because it shares our cold-start framing and accepts outcome-like
targets through validators. We could not run it head-to-head: it
executes against a hosted NVIDIA inference service behind an API key
rather than as a local library, so it does not fit the isolated,
version-pinned environment the rest of the leaderboard uses. We decline
to fabricate a number for it. The comparison is not vacuous, though,
because two results in this paper bound its category. NeMo is an
LLM-stochastic, validator-guided generator: its validators check and
retry, they do not guarantee, so it cannot produce an exact aggregate or
bitwise determinism by construction. The per-period conditioned steelman
is the strongest form of an approximating method we could engineer, and
it lands at AME 0.189, never 0 (this row). And E12 is, in effect, a
direct measurement of an LLM-stochastic specification-to-data pipeline
of exactly NeMo's kind: two different language models behind our own
front-end each cover the domains best-effort and fail on different
cases, while the exactness appears only once a valid spec reaches the
deterministic engine. So the structural claim against NeMo, that an
LLM-validator loop approximates but cannot exactly satisfy an analytical
outcome, is supported by measurements we did run, not by a missing row.
A direct NeMo comparison, on a machine with NVIDIA service access,
remains worth adding.

Two further points keep the result inside its evidence. On the metric's
validity (review F2), the monthly targets here are not near-uniform:
their coefficient of variation is 0.30, the largest month is 2.2 times
the median, and the binning (\texttt{HouseAge\ mod\ 12}) is essentially
uncorrelated with value (\texttt{r\ =\ -0.03}). So the miss is a real
failure to match per-bin structure, confirmed month by month. For May
the real target is 7,103 while a fitted GaussianCopula sums to 3,318,
under half. These figures are reproduced in
\texttt{research/\allowbreak{}specbench/\allowbreak{}verify\_\allowbreak{}california\_\allowbreak{}bins.py}
(\texttt{results\_\allowbreak{}california\_\allowbreak{}bins.csv}). The miss is not an artifact of
a degenerate target. And on our own side (review M11), this schema
carries no domain prior, so the engine hits the aggregate exactly while
its per-row marginal is a supplied default. Misata and NaiveRescale are
both ``right aggregate, generic marginal'' here, and we claim no
marginal-realism advantage over a rescale on non-curated data. The
defensible real-data claim is precisely the conformance capability gap
against imitation, and nothing more.

\textbf{Reference-mode, relational depth (review M13).} To test the
relational claim against a purpose-built relational synthesizer, we ran
two depths against SDV's HMA: a two-table parent-to-child case
(customers to orders) and a three-table hierarchy (regions to stores to
sales, two foreign-key edges), each with an outcome target on the child
metric.

\begin{longtable}[]{@{}llllll@{}}
\toprule
Task & Generator & input & AME (3 seeds) & FIVR & DET\tabularnewline
\midrule
\endhead
2-table & Misata (ours) & schema & \ensuremath{\approx}0
(\ensuremath{\le} 1.3 \ensuremath{\times} 10\textsuperscript{-7}) & 0 &
1\tabularnewline
2-table & SDV HMA (relational) & data & 0.776 & 0 & 1\tabularnewline
3-table & Misata (ours) & schema & \ensuremath{\approx}0
(\ensuremath{\le} 1.3 \ensuremath{\times} 10\textsuperscript{-7}) & 0 (2
edges) & 1\tabularnewline
3-table & SDV HMA (relational) & data & 0.698 & 0 (2 edges) &
1\tabularnewline
\bottomrule
\end{longtable}

The honest relational reading is that integrity does not separate the
two methods. HMA preserves every foreign key by construction, including
both edges of the three-level hierarchy, so its FIVR is 0 and ours is 0.
What separates them is conformance. HMA, trained on the real child
table, has no mechanism to ingest an outcome target, so it misses the
declared monthly aggregate by 78 percent at two tables and 70 percent at
three. The engine hits the aggregate at the rounding floor (at most
\texttt{1.3\ x\ 10\^{}\{-7\}} relative, sub-cent) with FIVR = 0 at both
depths. The relational contribution is therefore integrity \emph{jointly
with} outcome conformance, not a claim of better integrity than a
relational synthesizer that already ties us at 0. The run is
\texttt{research/\allowbreak{}specbench/\allowbreak{}run\_\allowbreak{}multitable.py}, with results in
\texttt{results\_\allowbreak{}multitable.csv}.

\textbf{Reference-mode, controlled check.} On a synthetic source table
with a clean ramp (10 seeds), used only to isolate behavior on a
known-smooth curve:

\begin{longtable}[]{@{}llllll@{}}
\toprule
Generator & input & CSC & AME (10 seeds) & DET & fit+sample
(s)\tabularnewline
\midrule
\endhead
Misata (ours) & schema & 1 & 0 & 1 & 0.04\tabularnewline
SDV GaussianCopula & data & 0 & 0.213 \ensuremath{\pm} 0.000 & 1 &
0.82\tabularnewline
SDV HMA & data & 0 & 0.213 \ensuremath{\pm} 0.000 & 1 &
1.54\tabularnewline
SDV CTGAN & data & 0 & 0.569 \ensuremath{\pm} 0.436 & 1 &
48.8\tabularnewline
Faker & schema & 1 & 0.895 \ensuremath{\pm} 0.005 & 1 &
0.00\tabularnewline
\bottomrule
\end{longtable}

CTGAN's 0.569 \ensuremath{\pm} 0.436 is the telling number. The standard
deviation is most of the mean, which is what an uncontrolled process
looks like: sometimes it nearly hits the curve, sometimes it badly
misses, because there is no targeting mechanism to hold it in place. One
correction belongs here. An earlier single-seed draft reported CTGAN as
``non-deterministic.'' Once SDV is seeded on both torch and numpy, all
SDV synthesizers are deterministic (DET = 1), and that earlier claim was
an artifact of an unseeded run, now retracted. We also concede the
obvious: where real data exists and the goal is fidelity to it, learned
methods lead. That is a different objective from the one measured here.

\hypertarget{exact-aggregates-cost-almost-no-shape-distortion-e6-proposition-4}{%
\subsection{6.3 Exact aggregates cost almost no shape distortion (E6,
Proposition
4)}\label{exact-aggregates-cost-almost-no-shape-distortion-e6-proposition-4}}

E6 is the falsifiable form of the scale-invariance claim, and it carries
a correction we keep in the text on purpose. An earlier draft reported a
``condensation frontier'': a 21.7\ensuremath{\times} rise in marginal
distortion as the tail grew heavier. Adding the proper unconstrained
control showed that rise to be a Beta-against-lognormal family mismatch
plus a finite-sample 1-Wasserstein bias, not an effect of the aggregate
constraint. That claim is retracted. The corrected experiment (10 seeds)
holds the aggregate exact and compares the engine's constrained sample
to an unconstrained i.i.d. draw from the \emph{same} family across
tail-heaviness, sweeping the coefficient of variation from 0.35 to 2.9.

The two distributions track each other throughout. The distortion gap
(constrained minus unconstrained) stayed negative across the whole
sweep, from about \ensuremath{-}0.01 at the lightest tail to about
\ensuremath{-}0.10 at the heaviest, meaning the constrained sample sits
as close to the target as the unconstrained control or slightly closer.
The aggregate stayed exact the whole way, with a largest period sum
error of \texttt{1.5\ x\ 10\^{}\{-11\}}. This is what Proposition 4
predicts. The engine fixes the per-row shape (a Dirichlet, equivalently
Gamma, family at concentration \texttt{alpha}) and lets the scale absorb
the constraint, so there is no fixed marginal to collapse. Condensation,
which afflicts conditional sampling under a \emph{fixed marginal}, is
avoided by construction rather than merely deferred. The figure is
\texttt{research/\allowbreak{}specbench/\allowbreak{}scale\_\allowbreak{}invariance.png}; its data are in
\texttt{scale\_\allowbreak{}invariance\_\allowbreak{}curve.csv} and
\texttt{scale\_\allowbreak{}invariance\_\allowbreak{}summary.csv}. We lose a dramatic frontier
plot and gain a smaller, correct one. We take that trade.

\hypertarget{conformance-beyond-temporal-sums-e7}{%
\subsection{6.4 Conformance beyond temporal sums
(E7)}\label{conformance-beyond-temporal-sums-e7}}

Outcome conformance is not limited to aggregate curves. We declared a
churn-rate target and a plan-distribution (group-share) target, and
measured Rate-Conformance Error (RCE = \textbar declared \ensuremath{-}
realized\textbar) and Group-Distribution Conformance (GDC, the
total-variation distance between declared and realized shares). Across
declared rates \{0.05, 0.10, 0.20, 0.35\} the engine held RCE under
0.004 (realized 0.051, 0.101, 0.201, and 0.347 against the four
targets). For a declared 0.60 / 0.30 / 0.10 plan split it realized 0.600
/ 0.299 / 0.101, a GDC of about 0.001. An uncontrolled draw with no
target channel, by contrast, sat at RCE \ensuremath{\approx} 0.30
against the same 0.20 churn target. The contribution is outcome
conformance in general, not a single curve-fitting trick. The demo is
\texttt{research/\allowbreak{}specbench/\allowbreak{}demo\_\allowbreak{}rate\_\allowbreak{}group.py}.

\hypertarget{the-output-is-a-usable-database-e8}{%
\subsection{6.5 The output is a usable database
(E8)}\label{the-output-is-a-usable-database-e8}}

Intrinsic metrics aside, a developer has to be able to query the result.
From the one-line spec ``an ecommerce store with 2000 customers, revenue
\$50k in January rising to \$200k in December,'' we generated the
dataset, loaded it into a real SQLite database, and ran the SQL a BI
tool or a test suite would run. A \texttt{LEFT\ JOIN} orphan check
returned 0 dangling foreign keys. The in-database monthly revenue rollup
returned \$50,000 in January and \$200,000 in December, so the declared
outcome holds in SQL, not only inside the generator. A
\texttt{WHERE\ amount\ \textless{}=\ 0} assertion returned 0 rows. One
sentence in, a queryable and outcome-correct test database out. The
script is \texttt{research/\allowbreak{}specbench/\allowbreak{}case\_\allowbreak{}study.py}.

\hypertarget{cost-and-the-absence-of-an-infeasible-regime-e9}{%
\subsection{6.6 Cost, and the absence of an infeasible regime
(E9)}\label{cost-and-the-absence-of-an-infeasible-regime-e9}}

Closed-form generation is training-free and scales close to linearly. On
a comparable single-table workload it ran from 0.06 seconds at 1,000
rows to 0.34 seconds at 50,000 rows, against SDV GaussianCopula's
fit-plus-sample time of 0.56 to 2.40 seconds, a speedup of roughly
7\ensuremath{\times} to 11\ensuremath{\times} (and the off-the-shelf
copula still has no target interface). The aggregate-conformance problem
also has no infeasible regime. By Proposition 1 the exact sum is
reachable for any row count \texttt{n\ \textgreater{}=\ 1}, so pushing
targets to extremes (a \$1M month over 10 rows, or a \$0.02 month over
10 rows) still yields the exact sum; the second case simply leaves some
rows at 0. There is no silent-miss failure mode to detect because there
is no miss. The script is \texttt{research/\allowbreak{}specbench/\allowbreak{}throughput.py}.

\hypertarget{a-task-the-reference-method-fails-e10}{%
\subsection{6.7 A task the reference method fails
(E10)}\label{a-task-the-reference-method-fails-e10}}

A benchmark its own author always wins is worth little, so SpecBench
includes P\ensuremath{\star}: hit an exact monthly sum \emph{and} match
a specified external heavy-tailed marginal (a Pareto target). The engine
hits the sum and fails the marginal. Across 10 seeds the period sum
error was 0 at cent precision, while the normalized 1-Wasserstein
distance to a Pareto (shape 1.5) target averaged 1.06 (standard
deviation 0.14), far from the 0 a match would give. The magnitude tracks
the chosen tail: a heavier target pushes it higher. The failure is the
one Proposition 4 predicts: the engine fixes a Gamma-family shape and
cannot reproduce an arbitrary external marginal. This is the open,
condensation-bounded problem of Section 3.4, the fixed-marginal
exact-aggregate case we explicitly do not solve. SpecBench reports it as
a Misata failure, which is the point. The suite contains tasks the
proposing method loses, and the boundary of the contribution is measured
rather than hidden. The check is \texttt{research/\allowbreak{}specbench/\allowbreak{}pstar.py},
with results in \texttt{results\_\allowbreak{}pstar.csv}.

\hypertarget{the-frontier-mapped-the-constraint-is-free-the-shape-family-is-the-cost-e11}{%
\subsection{6.8 The frontier, mapped: the constraint is free, the shape
family is the cost
(E11)}\label{the-frontier-mapped-the-constraint-is-free-the-shape-family-is-the-cost-e11}}

P\ensuremath{\star} is open, but we can say precisely what makes it
hard, and the answer is not what an earlier draft assumed. The natural
worry is that forcing an exact sum is what stops the engine from
matching an external marginal. E11 shows that is false. We sweep a
target marginal family across tail-heaviness (lognormal over its
log-scale, Pareto over its tail index), match the engine's marginal CV
to the target's, and then decompose the miss with the control whose
absence produced the retracted frontier. For each target we measure
three normalized 1-Wasserstein distances to a large target sample: the
engine's exact-sum output, an unconstrained i.i.d. draw from the same
Gamma shape and mean with no sum constraint, and a second target sample
(the finite-sample estimator floor).

\begin{longtable}[]{@{}llllll@{}}
\toprule
\begin{minipage}[b]{0.14\columnwidth}\raggedright
Target family\strut
\end{minipage} & \begin{minipage}[b]{0.14\columnwidth}\raggedright
CV of target\strut
\end{minipage} & \begin{minipage}[b]{0.14\columnwidth}\raggedright
MD, exact-sum engine\strut
\end{minipage} & \begin{minipage}[b]{0.14\columnwidth}\raggedright
MD, unconstrained Gamma\strut
\end{minipage} & \begin{minipage}[b]{0.14\columnwidth}\raggedright
constraint cost\strut
\end{minipage} & \begin{minipage}[b]{0.14\columnwidth}\raggedright
shape-family gap\strut
\end{minipage}\tabularnewline
\midrule
\endhead
\begin{minipage}[t]{0.14\columnwidth}\raggedright
lognormal (\ensuremath{\sigma}=0.3)\strut
\end{minipage} & \begin{minipage}[t]{0.14\columnwidth}\raggedright
0.31\strut
\end{minipage} & \begin{minipage}[t]{0.14\columnwidth}\raggedright
0.040\strut
\end{minipage} & \begin{minipage}[t]{0.14\columnwidth}\raggedright
0.040\strut
\end{minipage} & \begin{minipage}[t]{0.14\columnwidth}\raggedright
\ensuremath{-}0.000\strut
\end{minipage} & \begin{minipage}[t]{0.14\columnwidth}\raggedright
0.022\strut
\end{minipage}\tabularnewline
\begin{minipage}[t]{0.14\columnwidth}\raggedright
lognormal (\ensuremath{\sigma}=0.7)\strut
\end{minipage} & \begin{minipage}[t]{0.14\columnwidth}\raggedright
0.80\strut
\end{minipage} & \begin{minipage}[t]{0.14\columnwidth}\raggedright
0.115\strut
\end{minipage} & \begin{minipage}[t]{0.14\columnwidth}\raggedright
0.114\strut
\end{minipage} & \begin{minipage}[t]{0.14\columnwidth}\raggedright
+0.001\strut
\end{minipage} & \begin{minipage}[t]{0.14\columnwidth}\raggedright
0.091\strut
\end{minipage}\tabularnewline
\begin{minipage}[t]{0.14\columnwidth}\raggedright
lognormal (\ensuremath{\sigma}=1.1)\strut
\end{minipage} & \begin{minipage}[t]{0.14\columnwidth}\raggedright
1.53\strut
\end{minipage} & \begin{minipage}[t]{0.14\columnwidth}\raggedright
0.263\strut
\end{minipage} & \begin{minipage}[t]{0.14\columnwidth}\raggedright
0.260\strut
\end{minipage} & \begin{minipage}[t]{0.14\columnwidth}\raggedright
+0.003\strut
\end{minipage} & \begin{minipage}[t]{0.14\columnwidth}\raggedright
0.222\strut
\end{minipage}\tabularnewline
\begin{minipage}[t]{0.14\columnwidth}\raggedright
lognormal (\ensuremath{\sigma}=1.3)\strut
\end{minipage} & \begin{minipage}[t]{0.14\columnwidth}\raggedright
2.10\strut
\end{minipage} & \begin{minipage}[t]{0.14\columnwidth}\raggedright
0.385\strut
\end{minipage} & \begin{minipage}[t]{0.14\columnwidth}\raggedright
0.380\strut
\end{minipage} & \begin{minipage}[t]{0.14\columnwidth}\raggedright
+0.006\strut
\end{minipage} & \begin{minipage}[t]{0.14\columnwidth}\raggedright
0.324\strut
\end{minipage}\tabularnewline
\begin{minipage}[t]{0.14\columnwidth}\raggedright
Pareto (b=4.0)\strut
\end{minipage} & \begin{minipage}[t]{0.14\columnwidth}\raggedright
0.35\strut
\end{minipage} & \begin{minipage}[t]{0.14\columnwidth}\raggedright
0.491\strut
\end{minipage} & \begin{minipage}[t]{0.14\columnwidth}\raggedright
0.491\strut
\end{minipage} & \begin{minipage}[t]{0.14\columnwidth}\raggedright
+0.001\strut
\end{minipage} & \begin{minipage}[t]{0.14\columnwidth}\raggedright
0.460\strut
\end{minipage}\tabularnewline
\begin{minipage}[t]{0.14\columnwidth}\raggedright
Pareto (b=2.5)\strut
\end{minipage} & \begin{minipage}[t]{0.14\columnwidth}\raggedright
0.89\strut
\end{minipage} & \begin{minipage}[t]{0.14\columnwidth}\raggedright
0.944\strut
\end{minipage} & \begin{minipage}[t]{0.14\columnwidth}\raggedright
0.941\strut
\end{minipage} & \begin{minipage}[t]{0.14\columnwidth}\raggedright
+0.003\strut
\end{minipage} & \begin{minipage}[t]{0.14\columnwidth}\raggedright
0.889\strut
\end{minipage}\tabularnewline
\bottomrule
\end{longtable}

Two readings fall out, and both hold across the whole grid (10 seeds).
First, the exact-sum constraint is nearly free: the constraint cost, the
distance the engine's output sits beyond an unconstrained same-family
draw, never exceeds 0.006 anywhere from light to heavy tails, while the
period sum stays exact (largest sum error
\texttt{5.8\ x\ 10\^{}\{-11\}}). This is Proposition 0 made visible
against an external target: conditioning on the sum perturbs the per-row
law by only \texttt{O(1/\allowbreak{}U)} in total variation, so the engine's
exact-sum marginal and the unconstrained same-family marginal sit at
essentially the same distance from any \texttt{F}. The constraint cost
we measure is that \texttt{O(1/\allowbreak{}U)} term. Second, the miss that does
exist is shape-family mismatch: the gap above the finite-sample floor
runs from 0.02 for a near-Gaussian lognormal to 0.89 for a heavy Pareto,
and an unconstrained Gamma draw incurs the same gap. So enforcing
exactness is not the obstacle. The obstacle is that the engine speaks
one shape family, and an arbitrary target speaks another. This relocates
P\ensuremath{\star} precisely: it is a problem of widening the
closed-form shape family under an exact sum, not of paying for the sum.
It also yields a practical envelope. For moderate-dispersion targets (CV
up to about 0.5) the engine approximates an external marginal to within
an MD of 0.07 while hitting the sum exactly, and the approximation
degrades smoothly and predictably with tail-heaviness. Figure
\texttt{research/\allowbreak{}specbench/\allowbreak{}pstar\_\allowbreak{}frontier.png} shows the two readings
at a glance: the exact-sum curve sits on top of the unconstrained curve
(the constraint is free) while both rise above the floor (the
shape-family gap). The sweep is
\texttt{research/\allowbreak{}specbench/\allowbreak{}pstar\_\allowbreak{}frontier.py}, with results in
\texttt{results\_\allowbreak{}pstar\_\allowbreak{}frontier.csv}.

\begin{figure}[htbp]
\centering
\includegraphics[width=0.92\textwidth]{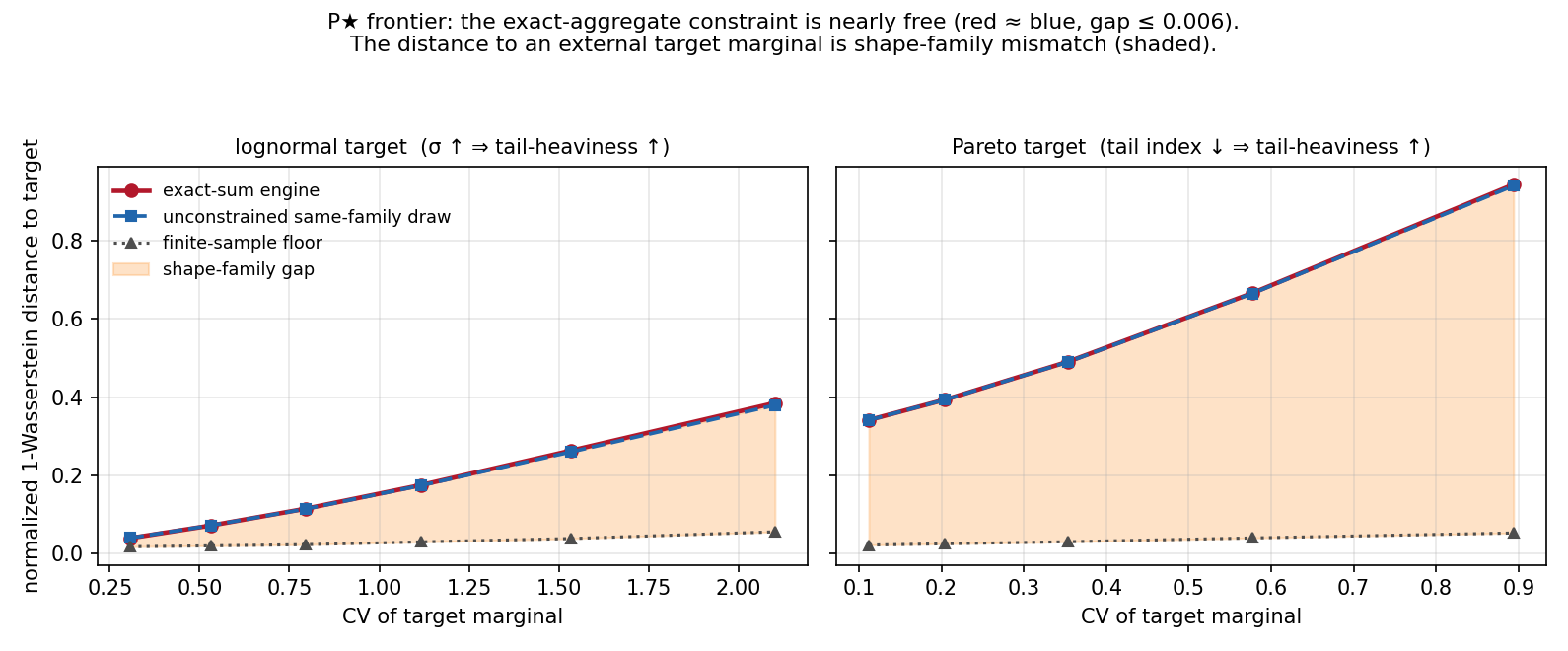}
\caption{The P\ensuremath{\star} frontier across target marginals. The exact-sum engine (solid) sits on the unconstrained same-family draw (dashed); enforcing the exact aggregate adds at most 0.006 in normalized 1-Wasserstein distance. The shaded gap to the target is shape-family mismatch.}
\label{fig:pstar}
\end{figure}

\hypertarget{suite-level-natural-language-conformance-across-18-domains-e12}{%
\subsection{6.9 Suite-level natural-language conformance across 18
domains
(E12)}\label{suite-level-natural-language-conformance-across-18-domains-e12}}

The leaderboard tasks exercise three curated domains through the
natural-language path. E12 asks the breadth question the benchmark
needs: state an outcome curve in one English sentence across all 18
domains (a January anchor rising to a December anchor) and check whether
the natural-language path yields a relational dataset that hits the
declared curve. We run three parsers behind the same engine: the
rule-based StoryParser that ships with the system and needs no API, and
two language-model front-ends, an open-weights model (Llama-3.3-70b via
Groq) and a recent OpenAI flagship (a GPT-5.3-class chat model). Per
domain we record whether the parser detected the curve, whether it read
the stated anchors, and then the engine's AME, FIVR, and determinism on
the generated tables.

\begin{longtable}[]{@{}llll@{}}
\toprule
\begin{minipage}[b]{0.22\columnwidth}\raggedright
Parser\strut
\end{minipage} & \begin{minipage}[b]{0.22\columnwidth}\raggedright
curve detected\strut
\end{minipage} & \begin{minipage}[b]{0.22\columnwidth}\raggedright
anchors correct\strut
\end{minipage} & \begin{minipage}[b]{0.22\columnwidth}\raggedright
conformant (AME \ensuremath{\approx} 0, FIVR = 0, DET = 1)\strut
\end{minipage}\tabularnewline
\midrule
\endhead
\begin{minipage}[t]{0.22\columnwidth}\raggedright
rule-based StoryParser (no API)\strut
\end{minipage} & \begin{minipage}[t]{0.22\columnwidth}\raggedright
8 / 18\strut
\end{minipage} & \begin{minipage}[t]{0.22\columnwidth}\raggedright
8 / 8\strut
\end{minipage} & \begin{minipage}[t]{0.22\columnwidth}\raggedright
8 / 8 (AME \ensuremath{\le} 4.3 \ensuremath{\times}
10\textsuperscript{-6})\strut
\end{minipage}\tabularnewline
\begin{minipage}[t]{0.22\columnwidth}\raggedright
Llama-3.3-70b\strut
\end{minipage} & \begin{minipage}[t]{0.22\columnwidth}\raggedright
18 / 18\strut
\end{minipage} & \begin{minipage}[t]{0.22\columnwidth}\raggedright
18 / 18\strut
\end{minipage} & \begin{minipage}[t]{0.22\columnwidth}\raggedright
18 / 18\strut
\end{minipage}\tabularnewline
\begin{minipage}[t]{0.22\columnwidth}\raggedright
GPT-5.3-class\strut
\end{minipage} & \begin{minipage}[t]{0.22\columnwidth}\raggedright
18 / 18\strut
\end{minipage} & \begin{minipage}[t]{0.22\columnwidth}\raggedright
18 / 18\strut
\end{minipage} & \begin{minipage}[t]{0.22\columnwidth}\raggedright
18 / 18\strut
\end{minipage}\tabularnewline
\bottomrule
\end{longtable}

The rule-based parser is reliable but narrow: on the 8 domains it
recognizes, anchor extraction is perfect and the engine hits the curve
to within \texttt{4.3\ x\ 10\^{}\{-6\}}, with no foreign-key violations
and bitwise determinism, but it does not recognize the other 10. Both
language models widen the surface to all 18 domains and read every
anchor correctly.

The path to the two 18-of-18 rows is the part worth reporting, because
it is where a stronger model turned out not to be the fix. On the first
pass each language model failed exactly one domain, and not the same
one. Llama-3.3 typed the gaming curve's time column as an integer month;
the GPT-5.3-class model wrote a malformed dotted pointer
(\texttt{orders.customer\_\allowbreak{}id.order\_\allowbreak{}date}) for the marketplace curve.
Both are the same failure class, a time-column reference the schema
validator rejects, and the newer, stronger model did not remove it, it
relocated it to a different domain. We added a deterministic
normalization between the model and the engine that resolves a dotted
pointer to its leaf and repoints a non-date time column to a real date
column in the same table. A unit test
(\texttt{research/\allowbreak{}specbench/\allowbreak{}test\_\allowbreak{}timecol\_\allowbreak{}repair.py}) confirms it
resolves both patterns with no model call, and with it in place both
models reach 18 of 18 while the rule-based path is untouched.

That is the honest lesson, and it is stronger than a single leaderboard
number. Two different models fail in two different best-effort ways, and
reliability is restored by a deterministic component, not by a larger
model. The language layer widens coverage; it does not supply exactness.
Once a valid curve reaches the engine, AME = 0, FIVR = 0, and DET = 1
hold by construction, independent of which parser produced the curve and
independent of the model's run-to-run variance. The script is
\texttt{research/\allowbreak{}specbench/\allowbreak{}nl\_\allowbreak{}suite.py}, with results in
\texttt{results\_\allowbreak{}nl\_\allowbreak{}suite.csv}.

\hypertarget{related-work}{%
\section{7. Related work}\label{related-work}}

Six threads bound the contribution, and the full annotated bibliography
is in \texttt{research/\allowbreak{}05\_\allowbreak{}literature\_\allowbreak{}review.md}.

Imitation and learned synthesis is the paradigm we sit orthogonal to.
SDV \cite{patki2016sdv}, CTGAN \cite{xu2019ctgan}, TabDDPM
\cite{kotelnikov2023tabddpm}, and the relational deep models RelDiff
\cite{reldiff2025}, IRG \cite{irg2024}, and HCTGAN \cite{hctgan2024}
learn a distribution from real data and are judged on fidelity. They
require source data and provide no interface for an analytical-outcome
target, so on cold-start tasks they do not run. We concede fidelity to
them where real data exists.

Query-aware database test-data generation is our closest kin and the
lineage we extend. QAGen \cite{binnig2007qagen}, DataSynth
\cite{arasu2011datasynth}, projection-compliant generation
\cite{projectioncompliant2022}, and the DBMS-feature testing framework
\cite{lo2010framework} generate databases that satisfy declared
query-output cardinalities for optimizer and feature testing. We differ
on the target type (analytical outcomes rather than query
cardinalities), the interface (declarative outcome curves and, on
curated domains, natural language), the method (closed-form
conditional-sum sampling rather than constraint solving), and the
evaluation (conformance metrics rather than cardinality match and
runtime).

LLM cold-start generation is the live competitor. NeMo Data Designer
\cite{nemo_data_designer} generates from scratch or from a seed with
statistical samplers and language models, validators, and model-as-judge
scoring, and the 2025 to 2026 wave (RDDG \cite{rddg2026}, LLM-TabLogic
\cite{llmtablogic2026}, StructSynth \cite{structsynth2025}, FASTGEN
\cite{fastgen2025}) extends it. All of these are stochastic and
approximate: validators check and retry, they do not guarantee. Our
wedge is the property they structurally lack, which is exact and
deterministic and closed-form conformance, and which SpecBench measures.
We cannot run NeMo against a hosted service in our isolated environment,
but Section 6.2 bounds its category with measurements we did run: the
conditioned steelman caps the approximating class at AME 0.189, and E12
shows an LLM-stochastic spec-to-data pipeline is best-effort, with
exactness supplied only by the deterministic engine downstream.

Aggregate-consistent methods in official statistics are where our
engine's exactness lives. Temporal disaggregation (Denton
\cite{denton1971}, Chow and Lin \cite{chowlin1971}) and iterative
proportional fitting \cite{deming1940ipf} produce aggregate-consistent
series and contingency tables. Our single-series aggregate task is
mathematically a disaggregation, and Denton is a baseline. The
difference is that we generate a population of transaction rows per
period with a controlled marginal, not one value per sub-period, and we
do it across a relational schema.

The exact mathematics is classical and cited as such: Lukacs
proportion-sum independence \cite{lukacs1955gamma}, compositional data
analysis \cite{aitchison1986compositional}, controlled rounding
\cite{cox1987rounding}, and largest-remainder apportionment
\cite{balinski1982fair}. The boundary is condensation of conditioned
sums \cite{armendariz2011condensation,szavits2014condensation}, the
obstruction for the fixed-marginal problem the engine sidesteps,
alongside the constrained-sampling literature \cite{golchi2016scmc}. On
evaluation, recent fidelity and utility benchmarks
\cite{multidim_benchmark2025,syntheval2024,tabsyndex2022} are uniformly
reference-based; they confirm the gap rather than fill it, since none
measures conformance to a declared specification. We unify the
specification thread, supply the missing conformance benchmark, and
reuse and cite the rest.

\hypertarget{limitations}{%
\section{8. Limitations}\label{limitations}}

Exact aggregate satisfaction is per-column and per-period independent.
Joint targets across correlated metrics or across foreign-key joins are
not handled and are future work. The engine fixes a Gamma-family shape
(Proposition 4), so hitting an arbitrary externally specified marginal
under an exact aggregate (P\ensuremath{\star}, Section 6) is open and
condensation-bounded; the suite reports this as a task the method fails,
and E11 locates the obstacle in the shape family rather than the
exact-sum constraint. Widening the closed-form shape family under an
exact aggregate is the natural next step, and it is unsolved here. On
cold-start spec-mode, exact aggregation alone is trivial, since a
rescale ties it given a hand-built schema, and the contribution there is
conformance from a specification with zero data, not the aggregate
number itself. On a non-curated schema the engine has no domain prior,
so it hits the aggregate exactly while its per-row marginal is a
supplied default; we claim no marginal-realism advantage over a rescale
in that case, only the conformance capability gap against imitation.
Domain priors are curated rather than learned, so realism outside the
curated domains rests on user specification. The natural-language
front-end was evaluated with two language models, an open-weights model
and a recent OpenAI flagship. Their failures are model-level and do not
coincide: each missed a different single domain on a malformed
time-column pointer, which a deterministic bridge repair then resolved
for both. A systematic sweep across more models, and across the very
newest reasoning models we did not run end to end here for runtime
reasons, is future work. The guarantees do not depend on that choice,
since a better parser raises coverage, not exactness. Our primary
reference-mode result is on one real public dataset (California
Housing); broader external validity across more real schemas is the
priority addition. Where real data exists and fidelity is the goal, we
do not compete with learned methods.

\hypertarget{conclusion}{%
\section{9. Conclusion}\label{conclusion}}

Outcome-conformant relational synthesis is a real and common problem
that the imitation paradigm does not serve: off-the-shelf it cannot take
the target, and no sampler can hit an aggregate exactly. We gave the
exact-aggregate engine its correct account (Proposition 0 reveals it as
conditional-sum sampling of a Gamma population, and Proposition 4 shows
it preserves the marginal by fixing the shape and letting the scale
absorb the constraint, which is how it sidesteps the condensation
obstruction that bounds the fixed-marginal problem). We built the first
conformance benchmark, SpecBench, and released a closed-form, zero-data,
integrity-preserving reference system. On reference-mode it attains AME
= 0 from the spec where learned methods, trained on target-consistent
data, reach 0.21 (GaussianCopula and HMA) to 0.57 \ensuremath{\pm} 0.44
(CTGAN, effectively uncontrolled) on a synthetic ramp, and miss by 74 to
86 percent on real housing data, with even a per-period steelman
stalling at about 0.19. On cold-start spec-mode it is the only method
that produces a conformant relational dataset from a sentence with no
source data and no hand-built schema, and across 18 domains the
natural-language path yields conformant data while the exactness stays
in the engine rather than the parser. We also mapped the boundary the
method does not cross: enforcing the exact aggregate costs almost
nothing in marginal fidelity even against an external heavy-tailed
target, so the open problem is one of shape family, not of the
constraint. The contribution is unification, measurement, and honesty,
including the retraction of two claims that did not survive proper
controls, rather than a new theorem. That is what the area needed.

\hypertarget{appendix-a.-reproducibility}{%
\section{Appendix A.
Reproducibility}\label{appendix-a.-reproducibility}}

The proposition checks E1 to E4 run through \texttt{research/\allowbreak{}measure.py}
against the released engine, with no extra dependencies. The
cross-paradigm leaderboard E5 runs through
\texttt{python\ -m\ research.specbench.runner} in the isolated
environment (\texttt{requirements-specbench.txt}, SDV 1.37.0), writing
\texttt{research/\allowbreak{}specbench/\allowbreak{}results\_\allowbreak{}e5.csv}, with SDV seeded on torch
and numpy and DET measured over at least three same-seed regenerations.
The Proposition 4 figure E6 runs through
\texttt{research.specbench.scale\_\allowbreak{}invariance} and then
\texttt{plot\_\allowbreak{}scale\_\allowbreak{}invariance}, writing the curve and summary CSVs and
the figure. The P\ensuremath{\star} check runs through
\texttt{research/\allowbreak{}specbench/\allowbreak{}pstar.py} (\texttt{results\_\allowbreak{}pstar.csv}), and
the P\ensuremath{\star} frontier (E11) through
\texttt{research/\allowbreak{}specbench/\allowbreak{}pstar\_\allowbreak{}frontier.py}
(\texttt{results\_\allowbreak{}pstar\_\allowbreak{}frontier.csv}). The multi-table results run
through \texttt{research/\allowbreak{}specbench/\allowbreak{}run\_\allowbreak{}multitable.py}, and the
California per-bin validity check through
\texttt{research/\allowbreak{}specbench/\allowbreak{}verify\_\allowbreak{}california\_\allowbreak{}bins.py}. The 18-domain
natural-language suite (E12) runs through
\texttt{research/\allowbreak{}specbench/\allowbreak{}nl\_\allowbreak{}suite.py}
(\texttt{results\_\allowbreak{}nl\_\allowbreak{}suite.csv}); its rule-based arm needs nothing, its
Groq arm needs \texttt{GROQ\_\allowbreak{}API\_\allowbreak{}KEY}, and its OpenAI arm needs
\texttt{OPENAI\_\allowbreak{}API\_\allowbreak{}KEY} (set \texttt{OPENAI\_\allowbreak{}MODEL} to choose the
model). The deterministic time-column bridge repair has a no-API-call
test in \texttt{research/\allowbreak{}specbench/\allowbreak{}test\_\allowbreak{}timecol\_\allowbreak{}repair.py}. Oracles
are frozen in the task definitions, and no metric reads any generator's
internals.

One reproducibility caveat, which we measured rather than assumed: the
SDV-based numbers are sensitive to the numerical platform, not only to
package versions. The reported figures were produced on macOS (Apple
Silicon, Python 3.14, SDV 1.37.0, copulas 0.14.1, rdt 1.21.0). We re-ran
the copula-based rows on a second platform (Linux x86, Python 3.10) with
the SDV, copulas, and rdt versions matched to the originals. The
multi-table HMA figures reproduced exactly (0.776 and 0.698). On
California Housing, however, the GaussianCopula and HMA miss came out at
0.708 against the 0.739 reported here, and the conditioned steelman at
0.212 against 0.189, and matching the copula and rdt versions did not
close that gap. The residual difference is platform-level floating point
in the copula fit and sampling (the linear-algebra backend differs
between macOS-ARM and Linux-x86), so it cannot be removed by pinning
Python packages alone. The conformance gap holds in size and meaning
across both platforms (a miss of roughly 71 to 74 percent for
off-the-shelf imitation, and a steelman that approximates but never
reaches 0); only the second decimal moves. CTGAN was not re-run on the
second platform because it requires torch. The engine's own quantities
(AME, FIVR, DET) are exact by construction and are platform- and
stack-independent. For a camera-ready, the SDV rows should be
regenerated once on a single fixed platform, with that platform and
stack recorded alongside the numbers.

\hypertarget{appendix-b.-proofs}{%
\section{Appendix B. Proofs}\label{appendix-b.-proofs}}

Full proofs are in \texttt{research/\allowbreak{}01\_\allowbreak{}formalization.md}: Proposition 0
(the Gamma-conditional identity), Proposition 1 (aggregate exactness),
Proposition 2 (the closed-form marginal CV), Proposition 3 (clamp
distortion), and Proposition 4 (scale-invariance, condensation avoided).
The retracted ``condensation frontier'' conjecture and its confound are
documented in \texttt{research/\allowbreak{}06\_\allowbreak{}adversarial\_\allowbreak{}review.md} and the
Section 3.4 retraction note.

\bibliographystyle{abbrvnat}
\bibliography{references}

\begin{thebibliography}{31}
\providecommand{\natexlab}[1]{#1}
\providecommand{\url}[1]{\texttt{#1}}
\expandafter\ifx\csname urlstyle\endcsname\relax
  \providecommand{\doi}[1]{doi: #1}\else
  \providecommand{\doi}{doi: \begingroup \urlstyle{rm}\Url}\fi

\bibitem[{\AA}gren and {\'U}beda~Sosa(2024)]{hctgan2024}
W.~{\AA}gren and V.~{\'U}beda~Sosa.
\newblock Hierarchical conditional tabular {GAN} for multi-tabular synthetic
  data generation.
\newblock 2024.
\newblock Preprint, arXiv:2411.07009.

\bibitem[Aitchison(1986)]{aitchison1986compositional}
J.~Aitchison.
\newblock \emph{The Statistical Analysis of Compositional Data}.
\newblock Chapman \& Hall, London, 1986.
\newblock ISBN 0-412-28060-4.

\bibitem[Arasu et~al.(2011)Arasu, Kaushik, and Li]{arasu2011datasynth}
A.~Arasu, R.~Kaushik, and J.~Li.
\newblock {DataSynth}: Generating synthetic data using declarative constraints.
\newblock \emph{Proc. VLDB Endowment (PVLDB)}, 4\penalty0 (12), 2011.
\newblock \doi{10.14778/3402755.3402785}.

\bibitem[Armend{\'a}riz and Loulakis(2011)]{armendariz2011condensation}
I.~Armend{\'a}riz and M.~Loulakis.
\newblock Conditional distribution of heavy tailed random variables on large
  deviations of their sum.
\newblock \emph{Stochastic Processes and their Applications}, 121\penalty0
  (5):\penalty0 1138--1147, 2011.

\bibitem[Balinski and Young(1982)]{balinski1982fair}
M.~L. Balinski and H.~P. Young.
\newblock \emph{Fair Representation: Meeting the Ideal of One Man, One Vote}.
\newblock Yale University Press, 1982.
\newblock ISBN 0-300-02724-9.

\bibitem[Binnig et~al.(2007)Binnig, Kossmann, Lo, and
  {\"O}zsu]{binnig2007qagen}
C.~Binnig, D.~Kossmann, E.~Lo, and M.~T. {\"O}zsu.
\newblock {QAGen}: Generating query-aware test databases.
\newblock In \emph{Proc. ACM SIGMOD Int. Conf. on Management of Data}, pages
  341--352, Beijing, China, 2007.
\newblock \doi{10.1145/1247480.1247520}.

\bibitem[Chow and Lin(1971)]{chowlin1971}
G.~C. Chow and A.-l. Lin.
\newblock Best linear unbiased interpolation, distribution, and extrapolation
  of time series by related series.
\newblock \emph{The Review of Economics and Statistics}, 53\penalty0
  (4):\penalty0 372--375, 1971.

\bibitem[Chundawat et~al.(2022)Chundawat, Tarun, Mandal, Lahoti, and
  Narang]{tabsyndex2022}
V.~S. Chundawat, A.~K. Tarun, M.~Mandal, M.~Lahoti, and P.~Narang.
\newblock {TabSynDex}: A universal metric for robust evaluation of synthetic
  tabular data.
\newblock 2022.
\newblock Preprint, arXiv:2207.05295.

\bibitem[Cox(1987)]{cox1987rounding}
L.~H. Cox.
\newblock A constructive procedure for unbiased controlled rounding.
\newblock \emph{Journal of the American Statistical Association}, 82\penalty0
  (398):\penalty0 520--524, 1987.

\bibitem[{DataCebo}()]{sdgym}
{DataCebo}.
\newblock {SDGym} and {SDMetrics}.
\newblock software + documentation.

\bibitem[Deming and Stephan(1940)]{deming1940ipf}
W.~E. Deming and F.~F. Stephan.
\newblock On a least squares adjustment of a sampled frequency table when the
  expected marginal totals are known.
\newblock \emph{The Annals of Mathematical Statistics}, 11\penalty0
  (4):\penalty0 427--444, 1940.
\newblock \doi{10.1214/aoms/1177731829}.

\bibitem[Denton(1971)]{denton1971}
F.~T. Denton.
\newblock Adjustment of monthly or quarterly series to annual totals: An
  approach based on quadratic minimization.
\newblock \emph{Journal of the American Statistical Association}, 66\penalty0
  (333):\penalty0 99--102, 1971.

\bibitem[Esteban et~al.(2017)Esteban, Hyland, and R{\"a}tsch]{esteban2017tstr}
C.~Esteban, S.~L. Hyland, and G.~R{\"a}tsch.
\newblock Real-valued (medical) time series generation with recurrent
  conditional gans.
\newblock 2017.
\newblock Preprint, arXiv:1706.02633.

\bibitem[Golchi and Campbell(2016)]{golchi2016scmc}
S.~Golchi and D.~A. Campbell.
\newblock Sequentially constrained monte carlo.
\newblock \emph{Computational Statistics \& Data Analysis}, 97:\penalty0
  98--113, 2016.

\bibitem[Hudovernik et~al.(2025)Hudovernik, Xu, Shi, {\v{S}}ubelj, Ermon,
  {\v{S}}trumbelj, and Leskovec]{reldiff2025}
V.~Hudovernik, M.~Xu, J.~Shi, L.~{\v{S}}ubelj, S.~Ermon, E.~{\v{S}}trumbelj,
  and J.~Leskovec.
\newblock {RelDiff}: Relational data generative modeling with graph-based
  diffusion models.
\newblock 2025.
\newblock Preprint, arXiv:2506.00710.

\bibitem[Kotelnikov et~al.(2023)Kotelnikov, Baranchuk, Rubachev, and
  Babenko]{kotelnikov2023tabddpm}
A.~Kotelnikov, D.~Baranchuk, I.~Rubachev, and A.~Babenko.
\newblock {TabDDPM}: Modelling tabular data with diffusion models.
\newblock In \emph{Proc. 40th Int. Conf. on Machine Learning (ICML), PMLR},
  volume 202, pages 17564--17579, 2023.

\bibitem[Lautrup et~al.(2025)Lautrup, Hyrup, Zimek, and
  Schneider-Kamp]{syntheval2024}
A.~D. Lautrup, T.~Hyrup, A.~Zimek, and P.~Schneider-Kamp.
\newblock {SynthEval}: A framework for detailed utility and privacy evaluation
  of tabular synthetic data.
\newblock \emph{Data Mining and Knowledge Discovery}, 39\penalty0 (1), 2025.

\bibitem[Li et~al.(2023)Li, Zhao, Abdollahzadeh, Sikdar, and Tay]{irg2024}
J.~Li, Z.~Zhao, M.~Abdollahzadeh, B.~Sikdar, and Y.~C. Tay.
\newblock {IRG}: Modular synthetic relational database generation with complex
  relational schemas.
\newblock 2023.
\newblock \doi{10.1145/3770854.3780313}.
\newblock Preprint, arXiv:2312.15187.

\bibitem[Liu et~al.(2025)Liu, Zheng, and Zhang]{structsynth2025}
S.~Liu, Y.~Zheng, and Y.~Zhang.
\newblock {StructSynth}: Leveraging {LLMs} for structure-aware tabular data
  synthesis in low-data regimes.
\newblock 2025.
\newblock Preprint, arXiv:2508.02601.

\bibitem[Lo et~al.(2010)Lo, Binnig, Kossmann, {\"O}zsu, and
  Hon]{lo2010framework}
E.~Lo, C.~Binnig, D.~Kossmann, M.~T. {\"O}zsu, and W.-K. Hon.
\newblock A framework for testing {DBMS} features.
\newblock \emph{The VLDB Journal}, 19\penalty0 (2):\penalty0 203--230, 2010.
\newblock \doi{10.1007/s00778-009-0157-y}.

\bibitem[Long et~al.(2025)Long, Xu, and Brintrup]{llmtablogic2026}
Y.~Long, L.~Xu, and A.~Brintrup.
\newblock {LLM-TabLogic}: Preserving inter-column logical relationships in
  synthetic tabular data via prompt-guided latent diffusion.
\newblock 2025.
\newblock Preprint, arXiv:2503.02161.

\bibitem[Lukacs(1955)]{lukacs1955gamma}
E.~Lukacs.
\newblock A characterization of the gamma distribution.
\newblock \emph{The Annals of Mathematical Statistics}, 26\penalty0
  (2):\penalty0 319--324, 1955.

\bibitem[Nguyen et~al.(2025)Nguyen, Schafft, Hale, and Alfaro]{fastgen2025}
A.~Nguyen, S.~Schafft, N.~Hale, and J.~Alfaro.
\newblock {FASTGEN}: Fast and cost-effective synthetic tabular data generation
  with {LLMs}.
\newblock 2025.
\newblock Preprint, arXiv:2507.15839.

\bibitem[{NVIDIA (formerly Gretel)}(2025)]{nemo_data_designer}
{NVIDIA (formerly Gretel)}.
\newblock {NeMo} data designer, 2025.
\newblock Apache-2.0; software + documentation.

\bibitem[Patki et~al.(2016)Patki, Wedge, and Veeramachaneni]{patki2016sdv}
N.~Patki, R.~Wedge, and K.~Veeramachaneni.
\newblock The synthetic data vault.
\newblock In \emph{IEEE Int. Conf. on Data Science and Advanced Analytics
  (DSAA)}, pages 399--410, 2016.
\newblock \doi{10.1109/DSAA.2016.49}.

\bibitem[Sanghi et~al.(2022)Sanghi, Ahmed, and
  Haritsa]{projectioncompliant2022}
A.~Sanghi, S.~Ahmed, and J.~R. Haritsa.
\newblock Projection-compliant database generation.
\newblock \emph{Proc. VLDB Endowment (PVLDB)}, 15\penalty0 (5):\penalty0
  998--1010, 2022.
\newblock \doi{10.14778/3510397.3510398}.

\bibitem[Sidorenko et~al.(2025)Sidorenko, Platzer, Scriminaci, and
  Tiwald]{multidim_benchmark2025}
A.~Sidorenko, M.~Platzer, M.~Scriminaci, and P.~Tiwald.
\newblock Benchmarking synthetic tabular data: A multi-dimensional evaluation
  framework.
\newblock 2025.
\newblock Preprint, arXiv:2504.01908.

\bibitem[Szavits-Nossan et~al.(2014)Szavits-Nossan, Evans, and
  Majumdar]{szavits2014condensation}
J.~Szavits-Nossan, M.~R. Evans, and S.~N. Majumdar.
\newblock Condensation transition in joint large deviations of linear
  statistics.
\newblock \emph{Journal of Physics A: Mathematical and Theoretical},
  47\penalty0 (45):\penalty0 455004, 2014.

\bibitem[Xu et~al.(2019)Xu, Skoularidou, Cuesta-Infante, and
  Veeramachaneni]{xu2019ctgan}
L.~Xu, M.~Skoularidou, A.~Cuesta-Infante, and K.~Veeramachaneni.
\newblock Modeling tabular data using conditional {GAN}.
\newblock In \emph{Advances in Neural Information Processing Systems
  (NeurIPS)}, pages 7335--7345, 2019.

\bibitem[Yao et~al.(2025)Yao, Kr{\v{c}}o, Ganev, and
  de~Montjoye]{dcrdelusion2025}
Z.~Yao, N.~Kr{\v{c}}o, G.~Ganev, and Y.-A. de~Montjoye.
\newblock The {DCR} delusion: Measuring the privacy risk of synthetic data.
\newblock 2025.
\newblock Preprint, arXiv:2505.01524.

\bibitem[Zhang et~al.(2026)]{rddg2026}
C.~Zhang et~al.
\newblock Self-reinforcing controllable synthesis of rare relational data via
  bayesian calibration.
\newblock 2026.
\newblock Preprint, arXiv:2604.16817.

\end{thebibliography}
\end{document}